\documentclass[letterpaper, 10 pt, journal, twoside]{IEEEtran}  

\usepackage{comment}
\usepackage{graphicx}
\usepackage{hyperref}
\usepackage{cite}
\usepackage{amssymb}
\usepackage{amsmath}
\usepackage{mathtools}
\usepackage{mathastext}
\usepackage[utf8]{inputenc}

\usepackage[linesnumbered,ruled,noend,vlined]{algorithm2e}
\usepackage{algs}
\SetArgSty{textup}
\usepackage[dvipsnames]{xcolor}

\SetCommentSty{mycommfont}
\bstctlcite{IEEEexample:BSTcontrol}

\IEEEoverridecommandlockouts                              

\begin{document}

\title{
Single-query Path Planning Using \\ Sample-efficient Probability Informed Trees}

\author{Daniel Rakita$^{1}$, Bilge Mutlu, Michael Gleicher
\thanks{$^{1}$Authors are with the Department of Computer Sciences, University of Wisconsin--Madison, Madison 53706, USA }
\thanks{\tt\small [rakita|bilge|gleicher]@cs.wisc.edu}%
\thanks{This work was supported by a Microsoft Research PhD Fellowship, National Science Foundation award 1830242, and a NASA University Leadership Initiative (ULI) grant awarded to the UW-Madison and The Boeing Company (Cooperative Agreement \# 80NSSC19M0124).}%
}


\markboth{IEEE Robotics and Automation Letters. Preprint Version. Accepted Month, Year}
{FirstAuthorSurname \MakeLowercase{\textit{et al.}}: ShortTitle} 
\IEEEpeerreviewmaketitle

\maketitle
\thispagestyle{empty}
\pagestyle{empty}



\begin{abstract}
In this work, we present a novel sampling-based path planning method, called \textit{SPRINT}.  The method finds solutions for high dimensional path planning problems quickly and robustly.  Its efficiency comes from minimizing the number of collision check samples. This reduction in sampling relies on heuristics that predict the likelihood that samples will be useful in the search process. Specifically, heuristics (1) prioritize more promising search regions; (2) cull samples from local minima regions; and (3) steer the search away from previously observed collision states.  Empirical evaluations show that our method finds shorter or comparable-length solution paths in significantly less time than commonly used methods.  We demonstrate that these performance gains can be largely attributed to our approach to achieve sample efficiency.
\end{abstract}

\section{Introduction}
\label{sec:intro}

Solving high-dimensional path planning problems, such as for robot manipulator motion planning, remains a challenging and important problem.  Many aspects of this problem are often addressed with \textit{sampling-based} path planners, which sample a collision-check function in order to probe the configuration space and inform a search strategy.  While, in theory, many sampling-based planners are guaranteed to eventually find a solution if one exists, in practice, even state-of-the-art approaches are unable to consistently solve challenging problems in reasonable amounts of time as they require many costly collision-check samples \cite{hauser2015lazy}.  




In this work, we present a novel sampling-based path planning method, called \textit{SPRINT} (\textbf{S}ample-efficient \textbf{PR}obability \textbf{IN}formed \textbf{T}rees) that finds solutions for high dimensional planning problems significantly faster than many state-of-the-art approaches. Our method uses heuristics designed to minimize the total number of collision-check samples required to find solutions by modeling how likely regions of the search space will yield useful samples.  These heuristics, as part of newly proposed global and local tree searches, are used to (1) prioritize more promising search regions to foster a greedy first global search strategy, while still exploring broadly in the limit; (2) cull samples from local minima regions in the local search to avoid wasteful, unfruitful samples; and (3) steer the local search away from previously observed collision states toward search regions predicted to contain more useful free-space samples.  Our heuristics can be implemented with efficient information storage and retrieval methods that further accelerate the search.  Specifically, our method does not perform expensive nearest neighbor checking each time a node is added to the search graph structure.  


\begin{figure}[t!]
	\includegraphics[width=\columnwidth]{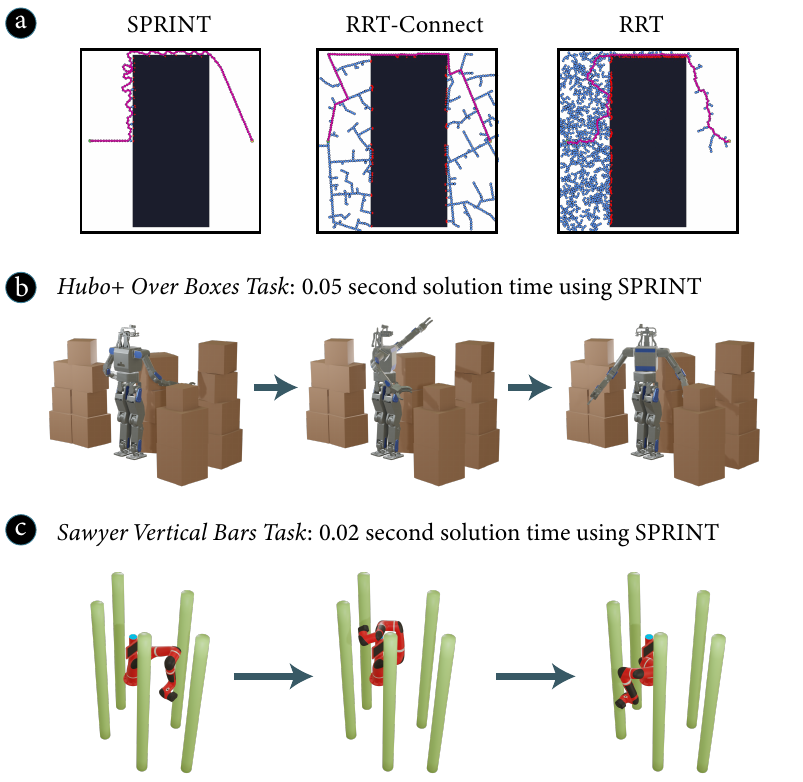}
	\caption{ (a) This 2D planning example shows the sample efficiency of the SPRINT planner, as many fewer collision-check samples were taken (left) compared to RRT-Connect (middle) and RRT (right).  (b--c) This sample efficiency also makes path planning more efficient in high dimensions, such as for the illustrated robot manipulator motion planning problems. }
	\label{fig:teaser}
	\vspace{0pt}
\end{figure}






We assessed the efficacy of our method by running a large testbed of simulation experiments and compared our method to numerous path planners on single-arm and bimanual robot manipulation problems (\S\ref{sec:evaluation}).  Our method found shorter or comparable-length solution paths in significantly less time than the alternatives.  We demonstrate that these performance advantages were, in large part, due to an increase in sample efficiency from our heuristics.  We conclude our work by discussing the implications of our results in robotics applications and beyond.  We provide open-source code for an implementation of our method.\footnote{\href{https://github.com/uwgraphics/lynx}{https://github.com/uwgraphics/lynx}}








\section{Related Works}
In general, there are three main paradigms for solving path planning problems: \textit{sampling-based}, \textit{local optimization}, and \textit{search-based}.  Sampling-based planners, many of which are extensions of Rapidly-Exploring Random Tree (RRT)\cite{lavalle1998rapidly}, Probabilistic Roadmap (PRM)\cite{kavraki1996probabilistic}, or Expansive Space Tree (EST)\cite{hsu1997path} methods, often use random configuration-space samples to bootstrap a broad search strategy.  These algorithms often have guarantees, such as probabilistic completeness or asymptotic optimality \cite{karaman2011sampling, janson2015fast}, though they often \textit{over-explore} the space with many unnecessary samples in order to achieve such guarantees.    


Our work shares particular similarities with \textit{informed-tree} sampling-based approaches that iteratively build search strategies that trade-off between exploration and exploitation based on information gathered at run-time \cite{burns2005toward, alterovitz2011rapidly, akgun2011sampling, gammell2014informed, gammell2015batch}.  Our work draws on these approaches, but differs in two main ways: (1) these approaches generally use linear local search segments between samples, whereas our work proposes a new local search designed to maneuver through narrow passages and avoid over-sampling; and (2) the search strategies for these approaches are typically more focused on achieving asymptotic optimality, whereas the search strategies for our method are more singularly focused on reaching a first feasible solution as quickly as possible.             

%


Local optimization-based planners, such as CHOMP\cite{ratliff2009chomp} or STOMP\cite{kalakrishnan2011stomp}, use non-linear optimization to iteratively transform a trajectory into a higher quality path.  These methods are designed to be greedier than sampling-based methods, though they often have fewer guarantees, strongly depend on the quality of the initial condition, and commonly do not converge on a feasible solution due to local minima.  

Search-based planners prioritize exploration based on a set \textit{heuristic}, such as best-first search, depth-first search, breadth-first search, or the A$^*$ heuristic\cite{aine2016multi}.  However, because these searches are often structured in a discrete, grid-like fashion, higher dimensional planning often scales intractably in terms of memory and run-time complexity.      

Our method draws on all of the above paradigms, and attempts to unite many of the discussed themes.  To illustrate, SPRINT also uses random samples to bootstrap its search and explore broadly, but local optimization and search-based concepts are used to optimize and prioritize certain parts of the search to increase the probability of sample usefulness.  We compare our method to other common approaches in our evaluation (\S\ref{sec:evaluation}) to assess whether this bridging of concepts is effective for path planning.

\section{Preliminaries}
\label{sec:background}


\subsection{Problem Statement}
Consider $\chi$ as a $d$-dimensional configuration space with states within $\chi$ denoted as $\mathbf{q}$. Suppose $\chi_{obs}$ is a subset of $\chi$ considered to be infeasible space (i.e., obstacles).  Feasible space can then be defined as the subset of $\chi$ \textit{not} in an obstacle region: $\chi_{free} \equiv \chi \setminus \chi_{obs}$.  A \textit{path} in $\chi$ is a continuous function $\Gamma[\mathbf{q}_a, \mathbf{q}_b] : [0,1] \rightarrow \chi$, where $\Gamma[\mathbf{q}_a, \mathbf{q}_b](0) = \mathbf{q}_a$, $\Gamma[\mathbf{q}_a, \mathbf{q}_b](1) = \mathbf{q}_b$, and all points along the path are in $\chi$.  The goal in path planning is to find a feasible, $C^0$ continuous path $\Gamma[\mathbf{q}_{init}, \mathbf{q}_{goal}]$ such that $\Gamma[\mathbf{q}_{init}, \mathbf{q}_{goal}](0)$ is a given start state, $\mathbf{q}_{init}$; $\Gamma[\mathbf{q}_{init}, \mathbf{q}_{goal}](1)$ is a given goal state, $\mathbf{q}_{goal}$; and $\Gamma[\mathbf{q}_{init}, \mathbf{q}_{goal}](u) \in \chi_{free}, \ \forall u \in [0,1]$.  






\subsection{Graph-based Planning Structure}
A common structure to solve the problem described above is a graph-based search \cite{lavalle1998rapidly, kavraki1996probabilistic, janson2015fast, karaman2011sampling, gammell2014informed, hsu1997path}.  This framework organizes the search into a set of collision-free edges, $E$, between pairs of nodes in a set $N$ to form a graph $G = (N, E)$.  The goal in these searches is to construct $G$ such that a feasible path $\Gamma[\mathbf{q}_{init}, \mathbf{q}_{goal}]$ is contained as some connected sequence of edge traversals between nodes within the graph.  The feasibility of each edge is checked prior to its addition to the graph using a collision-check function.  SPRINT interleaves global and local graph-based searches, overviewed in \S\ref{sec:structure}.

\subsection{Search-space Regions and Useful Samples}
\label{sec:search_space_regions_useful_samples}
The overall goal of SPRINT is to minimize the number of collision-check samples by incorporating probability heuristics that model how likely regions of the search space will yield useful samples.  


A \textit{search-space region} is a segment of the configuration space $\chi$ loosely thought of as states in the vicinity of the line segment between two specified end-points.  These regions are used to characterize and model where the planner has previously searched and where the planner could search in the future.  Their loose definition is sufficient for these use cases.  We denote a search-space region as $R[\mathbf{q}_a, \mathbf{q}_b]$, where $\mathbf{q}_a$ and $\mathbf{q}_b$ serve as the end-point markers of the region.    


Intuitively, a \textit{useful sample} is a sample that plays an integral role in constructing a final solution path.  More formally, we consider a sample to be $\delta-$useful if it achieves two criteria: (1) it is in free-space, $\chi_{free}$; and (2) it ultimately lies within a distance of $\delta$ of the final solution path, $\Gamma[\mathbf{q}_{init}, \mathbf{q}_{goal}]$.  Note that this definition implies that it is unknowable if a sample is $\delta-$useful until a solution path is found.  However, our method uses heuristics to predict the probability that a particular search-space region will contain at least one $\delta-$useful sample, which we will denote as $Pr(U_{\delta}(R[\mathbf{q}_a, \mathbf{q}_b]))$.  The following sections will detail the structure and functionalities of these probability heuristics. 


\section{Technical Overview}
This section outlines the search structure and overall strategy of the SPRINT method. Pseudocode of the SPRINT planning sub-processes can be found in Alg. \ref{alg:outerloop}--\ref{alg:collision_points}.   


\subsection{SPRINT Global planning Level}
\label{sec:structure}
The global planning level in SPRINT (Alg. \ref{alg:outerloop}) uses a tree-graph as a search structure, $T_g = (N_g, E_g)$, rooted at $\mathbf{q}_{init}$.  A set of sampled collision-free \textit{milestone points}, $M$, serve as intermediary goals for the global search to reach en route to $\mathbf{q}_{goal}$, reminiscent of FMT$^*$ \cite{janson2015fast}.  On each global planning level loop, the planner selects a best search-space region for the next local search, $R[\mathbf{q_n^*, q_m^*}]$ ($\mathbf{q}_n^* \in N_g$, $\mathbf{q}_m^* \in M$).  Milestones that are reached via local searches to form global edges in $E_g$ from already established nodes in $N_g$ are added as global nodes in $N_g$.  If the current set of milestone points does not foster a path to $\mathbf{q}_{goal}$, more milestone points are added to $M$, and the search proceeds.  This process iterates until $\mathbf{q}_{goal}$ is reached via $T_g$, provided a solution exists.

\algOuterLoop
\algLocalSearch


In general, the global planning level is intended to be a simple wrapper around the more sophisticated local search, outlined below.  While the global planning level achieves probabilistic completeness in the limit  (\S\ref{sec:analysis}), in practice, the default starting set of $50$ milestone points is sufficient to quickly solve all problems that we have ever tried. 



\subsection{SPRINT Local Planning Level}
\label{sec:local_planning_level}
The local planning level in SPRINT (Alg. \ref{alg:localsearch}) also uses a tree-graph as a search structure, $T_{\ell} = (N_{\ell}, E_{\ell})$.  The root of the local tree is the first boundary point of the search-space region selected by the global planning level, $\mathbf{q}_n^*$, and the goal state that the local tree is trying to reach is the second boundary point of this region, $\mathbf{q}_m^*$.  

The local planning level in SPRINT is a greedy, depth-first-search-like algorithm that uses heuristics to intelligently select branching directions, backtrack to fruitful parts of the search tree, and stop as early as possible when a solution is unlikely to be found.  The local search was designed to be particularly adept at steering around approximately convex-shaped obstacle sections and navigating through narrow passages (as seen in Figure \ref{fig:teaser}a).  While the local search has no guarantees and needs the global search to route around local minima regions, in practice, it is effective and solves many problems on its own, even in high dimensions. 



The local planning level progresses in three steps: (1) Select a node from the tree to extend from, $\mathbf{q}_x \in N_{\ell}$; (2) Decide if $\mathbf{q}_x$ is worthwhile to extend (if not, return to step 1); and (3) If $\mathbf{q}_x$ is worthwhile to extend, calculate a candidate node to extend toward, $\mathbf{q}_c^*$.  After $\mathbf{q}_c^*$ is calculated, it is checked for collision (Alg. \ref{alg:localsearch}, line 14).  If $\mathbf{q}_c^*$ is in free-space, it is added as a node to the tree, the algorithm assesses and stores information about how the search has improved or worsened given the addition of the new node, and it becomes the next $\mathbf{q}_x$ to extend (Alg. \ref{alg:localsearch}, lines 23--39).  If $\mathbf{q}_c^*$ is in collision, the collision-point is stored by the algorithm to help prevent the tree from colliding with the same region again and a next $\mathbf{q}_x$ is popped off of a stack of nodes to extend from, $N_{stack}$ (Alg. \ref{alg:localsearch}, lines 15--21).  These steps repeat until either the local goal is reached or $N_{stack}$ becomes empty.  


The heuristics used in the local search often rely on its ability to efficiently assess and characterize sub-trees within the tree.  Therefore, the local search tree stores information allowing fast responses to queries such as ``has the sub-tree rooted at node $\mathbf{q}$ recently progressed toward the goal $\mathbf{q}_m^*$?''; or ``what previously observed collision points are in close proximity to the sub-tree rooted at node $\mathbf{q}$?''.  To achieve these quick assessments, our method labels certain nodes in the local search tree as data \textit{checkpoints}, then stores and accesses information at these checkpoints throughout the search.  It would be highly inefficient for all nodes to be labeled as checkpoints. Thus, only the root node, $\mathbf{q}_n^*$, and nodes that have more than one child node are labeled as checkpoints, which reduces the overall number of checkpoints and structurally places them as roots of salient sub-trees where distinct branching decisions were made. 



\begin{figure}[t!]
	\includegraphics[width=\columnwidth]{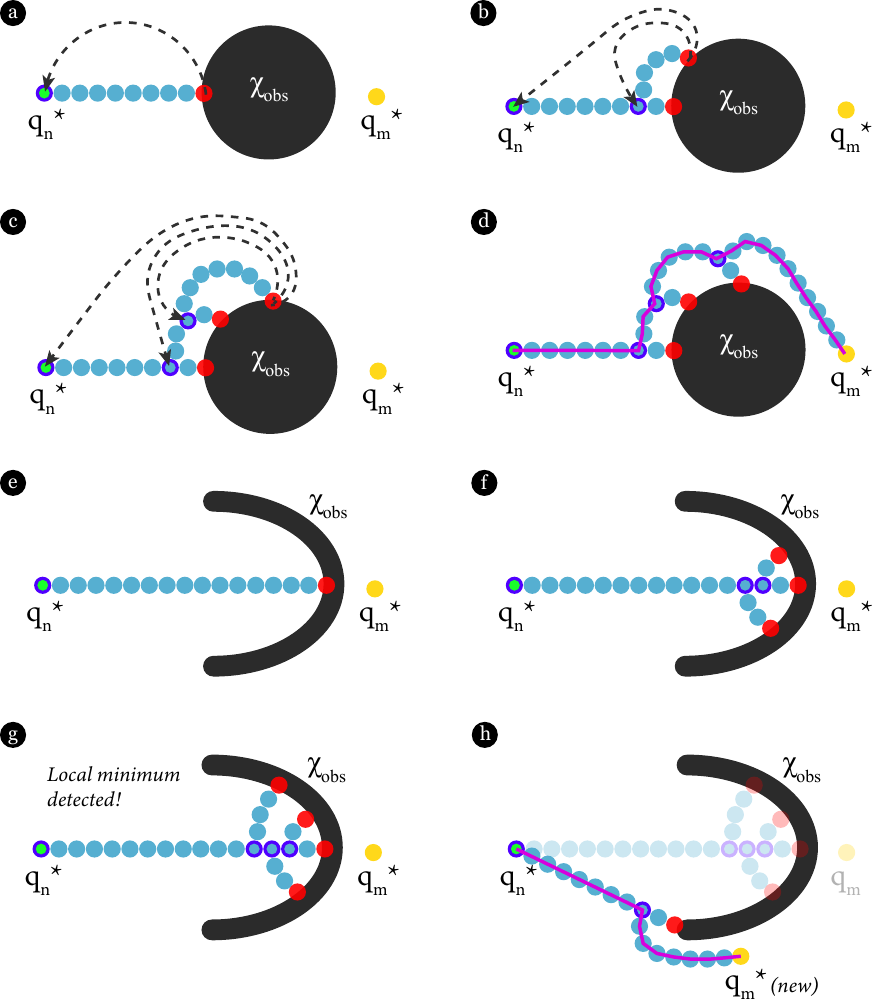}
	\caption{ (a) - (d) The detected collision states (red dots) get stored (dotted lines) in all checkpoint nodes (dots with purple outlines) along the path from the current extend node, $\mathbf{q}_x$ to the root, $\mathbf{q}_n^*$.  (e) - (f) Each added node passes its exploitation and exploration progress information back to its predecessor checkpoints until (g) the search decides that it is stuck in a local minimum trap based on lack of progress.  (h) A new search-space region is selected by the global search to route around the local minimum region. }
	\label{fig:backprop}
    \vspace{1pt}
\end{figure}

When pertinent information is processed during the search, \textit{e.g.}, a node is added or a collision point is detected, this information is back-propagated through the tree and stored at all checkpoint nodes that reside on the path from the current extend node to the local tree root, $\mathbf{q}_n^*$.  This process can be seen in Alg. \ref{alg:localsearch} lines 17 and 30--39 where information is updated in four hashmap data structures ($H_{exploit}$, $H_{explore}$, $H_{obs}$, $H_{num}$, explained more below) on checkpoint nodes $\mathbf{q} \in N_{checkpoints} \cap \Gamma[\mathbf{q}_n^*, \mathbf{q}_x]$.  This procedure ensures that checkpoints have access to information that has been processed ``downstream'' at newer, more distal parts of its sub-tree (illustrated in Figure \ref{fig:backprop}).

\section{Probability Heuristics}
In this work, our method uses three probability heuristics, one in the global planning level and two in the local planning level (highlighted in green in Alg. \ref{alg:outerloop} and \ref{alg:localsearch}).  Each heuristic tries to guide the search into search-space regions with a high probability of containing $\delta-$useful samples.  In this section, we describe each of these three heuristics.  



\algGlobalEdge

\subsection{Probability Heuristic 1} 

Probability Heuristic 1 is used by the global search to select which search-space region would be best to next attempt a local search.  This assessment is based on two criteria: (1) A good search-space region should have an end-point $\mathbf{q}_m^* \in M$ that gets closer to $\mathbf{q}_{goal}$.  Intuitively, global edges that make progress toward the goal are more likely to have useful samples; and (2) A good search-space region should be far from any regions that already fostered local searches that did not reach their respective goals.  The goal of this second criterion is to de-prioritize the search from repeating a local search in a region already estimated to be a local minimum trap (stored in the $R_{localmin}$ set in Alg. \ref{alg:outerloop}).   

\begin{figure}[t!]
	\includegraphics[width=\columnwidth]{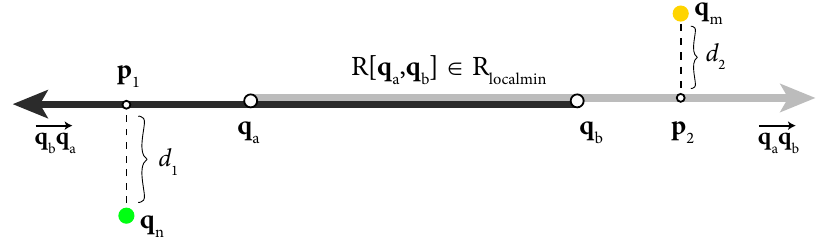}
	\caption{ Projection procedure of a search-space region $R[\mathbf{q}_n, \mathbf{q}_m]$ onto a region already deemed to be a local minimum trap, $R[\mathbf{q}_a, \mathbf{q}_b]$. }
	\label{fig:projection}
    \vspace{1pt}
\end{figure}

We model the two criteria outlined above as a mixture of two respective Gaussian functions:

\begin{equation}
\footnotesize
\begin{gathered}
\label{eq:probability_model1}
Pr_1(\ U_{\delta}(R[\mathbf{q}_g, \mathbf{q}_m]) \ | \ T_g\ ) = \\
\phi^{-1} * [(\textit{w}_1 * \textit{g}_1(\mathbf{q}_n, \mathbf{q}_m, T_g)) \ * \ (\textit{w}_2 * \textit{g}_2(\mathbf{q}_n, \mathbf{q}_m,  T_g))] \\
\end{gathered}
\end{equation}

These Gaussian functions can be seen in Alg. \ref{alg:global_edge}, lines 7 and 8--14.  The $\phi$ scalar is a normalization constant, and the $\textit{w}$ scalars signify weights that adjust the peaks of their Gaussian terms.  The default values for $\textit{w}_1$ and $\textit{w}_2$ were hand-tuned, though our experiments below show that performance is not overly sensitive to these parameters.  The $\textit{c}$ scalar is the standard deviation over these Gaussian functions, which we set as the Euclidean distance from start to goal.    

The $\textit{g}_1$ function uses Euclidean distances to achieve the first criterion discussed above.  Specifically, $\textit{x}_1$, the input variable of the $\textit{g}_1$ Gaussian function, is designed such that the closer a search-space region gets to $\mathbf{q}_{goal}$ compared to where it started, the higher the probability value will be from $\textit{g}_1$.  The $\textit{g}_2$ function achieves the second criterion by projecting all candidate search-space regions onto rays cast from the end-points of prior unfruitful search-space regions (stored in the set $R_{localmin}$) and deprioritizing candidate regions with end-points close to any of these projections (Alg. \ref{alg:global_edge}, lines 9--14, seen illustrated in Figure \ref{fig:projection}).  

\subsection{Probability Heuristic 2} 

Probability Heuristic 2 is used by the local search to decide whether a particular node in the local tree search,  $\mathbf{q}_{x}$, is worthwhile to extend.  This decision is made by checking information stored at all checkpoints along the path from $\mathbf{q}_x$ back to $\mathbf{q}_n^*$.  At a high level, $\mathbf{q}_x$ is deemed worthwhile to extend if either (1) all of the sub-trees rooted at checkpoints along the path from $\mathbf{q}_x$ back to $\mathbf{q}_n^*$ have shown recent progress in terms of getting closer to the goal, $\mathbf{q}_m^*$.  Throughout this work, we refer to this kind of progress as \textit{exploitation} (stored and tracked in the $H_{exploit}$ hashmap in Alg. \ref{alg:localsearch}); or (2) all of the sub-trees rooted at checkpoints along the path from $\mathbf{q}_x$ back to $\mathbf{q}_n^*$ have shown recent progress in terms of getting further from the tree root, $\mathbf{q}_n^*$.  Throughout this work, we refer to this kind of progress as \textit{exploration} (stored and tracked in the $H_{explore}$ hashmap in Alg. \ref{alg:localsearch}).  If the sub-trees that $\mathbf{q}_x$ is a part of show no sign of recent exploitation or exploration progress, the heuristic considers this region of the local tree trapped in a local minimum, and $\mathbf{q}_x$ is culled prior to extension.

We model Probability Heuristic 2 as the minimum of a set of probability functions:

\begin{equation}
\small
\begin{gathered}
\label{eq:probability_model2}
Pr_2( \ U_{\delta}(R[\mathbf{q}_x, \mathbf{q}_m^*])\ | \ T_{\ell} \ ) = \\
\min_{\mathbf{q} \in B}( \ \textit{g}(\mathbf{q}_x, \mathbf{q}, T_{\ell})  \ ), \ B \equiv N_{checkpoints} \cap \Gamma[\mathbf{q}_n^*, \mathbf{q}_x]
\end{gathered}
\end{equation}

The probability functions are modeled as Gaussian functions, constructed in Alg. \ref{alg:valid_node}, lines 4 -- 9.  Here, $\textit{x}_1$ and $\textit{x}_2$ are the number of collision-check samples since a node has made exploitation and exploration progress, respectively.  These values are stored and updated in the $H_{exploit}$ and $H_{explore}$ hashmaps in Alg. \ref{alg:localsearch}, lines 17 and 31--38.  We model this sub-tree assessment process by taking the minimum of $\textit{x}_1$ and $\textit{x}_2$ in Alg. \ref{alg:valid_node}, line 7.  Taking the minimum ensures that if either exploitation or exploration progress are occurring in the given sub-tree (\textit{i.e.}, either $\textit{x}_1$ or $\textit{x}_2$ are near 0), the sub-tree as a whole will still be considered promising.  This value is inputted into the Gaussian function in line 9 to output the approximate probability value $\textit{g} \in [0,1]$.  The standard-deviation $\textit{c}$ value is attenuated based on the number of nodes in a given sub-tree (stored and updated in the hashmap $H_{num}$ in Alg. \ref{alg:localsearch}) such that new sub-trees with few nodes do not fail too abruptly even if they do not immediately show exploitation or exploration progress.  In our prototype system, we use a probability cut-off of $\kappa = 0.3$, though our experiments below suggest that performance is not overly sensitive to this parameter selection.    

\algValidNode

\subsection{Probability Heuristic 3} 

Probability Heuristic 3 is used by the local search to calculate which search-space region the local tree should extend into from an extend node $\mathbf{q}_x$.  This region is computed based on three criteria: (1) A good search-space region should follow a straight line path with respect to the predecessor edge of $\mathbf{q}_x$.  This criterion prevents the local search from making drastic turns and unnecessarily turning back on its own path; (2) A good search-space region should get closer to the local goal, $\mathbf{q}_m^*$.  Intuitively, making progress toward the goal is estimated to be useful; and (3) A good search-space region should move away from previously observed collision points nearby (stored in the $H_{obs}$ hashmap in Alg. \ref{alg:localsearch}).  A search-space region that is pointing in the direction of a previously observed collision point is also likely to collide, and thus, is less likely to lead to a useful sample.

Instead of explicitly modeling a $Pr_3$ model, we implicitly define the model through its gradient with respect to a search-space region end-point, $\mathbf{q}_c$: $\frac{\partial Pr_3(\ U_{\delta}(R[\mathbf{q}_x, \mathbf{q}_c])\ )}{\partial \mathbf{q}_c}$.  Then, given this gradient, we perform gradient ascent to steer local edges in a way that would approximately optimize a $Pr_3$.  We model this gradient as proportional to the sum of three weighted sub-term gradients, each trying to achieve one of the three criteria from Probability Heuristic 3 above, respectively:

\algCommonFunctions
\algLocalEdge

\begin{equation}
\footnotesize
\begin{gathered}
\label{eq:probability_model3}
\frac{\partial Pr_3( \ U_{\delta}(R[\mathbf{q}_x, \mathbf{q}_c])) \ | \ T_{\ell} \ )}{\partial \mathbf{q}_c} \propto \textit{w}_1\frac{\partial \textit{g}_1}{\partial \mathbf{q}_c} + \textit{w}_2\frac{\partial \textit{g}_2}{\partial \mathbf{q}_c} + \textit{w}_3\frac{\partial \textit{g}_3}{\partial \mathbf{q}_c}
\end{gathered}
\end{equation}

The $\frac{\partial \textit{g}_1}{\partial \mathbf{q}_c}$ sub-term gradient pulls the search-space region in a straight line with respect to its predecessor edge.  Here, $\mathbf{q}_p$ denotes the predecessor node of the extend node, $\mathbf{q}_x$:

\begin{equation}
\footnotesize
\begin{gathered}
\label{eq:probability_model3_1}
\frac{\partial g_1}{\partial \mathbf{q}_c} = \frac{\mathbf{q}_x - \mathbf{q}_p}{||\mathbf{q}_x - \mathbf{q}_p||}
\end{gathered}
\end{equation}

The $\frac{\partial \textit{g}_2}{\partial \mathbf{q}_c}$ sub-term gradient pulls the search-space region toward the local goal, $\mathbf{q}_m^*$:

\begin{equation}
\footnotesize
\begin{gathered}
\label{eq:probability_model3_2}
\frac{\partial g_2}{\partial \mathbf{q}_c} = \psi_2 * \frac{\mathbf{q}_m^* - \mathbf{q}_c}{||\mathbf{q}_m^* - \mathbf{q}_c||} \\
\psi_2 = \textit{exp}(-||\mathbf{q}_m^* - \mathbf{q}_c||^2 \ / \ 4\lambda^2)+1
\end{gathered}
\end{equation}

Here, $\psi_2$ increases the pull from $\frac{\partial \textit{g}_2}{\partial \mathbf{q}_c}$ as the local tree approaches its goal which helps avoid overshoot when doing gradient ascent.  The $\lambda$ scalar is the fixed length of all local search edges. 

The $\frac{\partial \textit{g}_3}{\partial \mathbf{q}_c}$ sub-term gradient pushes the candidate edge away from a set of nearby, previously observed collision points $\mathbf{q}_{obs} \in N_{obs}$ (utility functions used in this sub-term gradient can be found in Alg. \ref{alg:common_functions}):


\begin{equation}
\footnotesize
\begin{gathered}
\label{eq:probability_model3_3}
\frac{\partial g_3}{\partial \mathbf{q}_c} = \frac{1}{|N_{obs}|}\sum_{\mathbf{q}_{obs} \in N_{obs}} \psi_{3,1} * \psi_{3,2} * \frac{ \texttt{proj}(\mathbf{q}_{obs}, R[\mathbf{q}_x, \mathbf{q}_c]) - \mathbf{q}_{obs}  }{ ||\texttt{proj}(\mathbf{q}_{obs}, R[\mathbf{q}_x, \mathbf{q}_c]) - \mathbf{q}_{obs}|| } \\
\psi_{3,1} = \texttt{hvs}(\texttt{proj\_scalar}(\mathbf{q}_{obs}, R[\mathbf{q}_x, \mathbf{q}_c])) \\ 
\psi_{3,2} = 5 *\textit{exp}(-||\texttt{proj}(\mathbf{q}_{obs}, R[\mathbf{q}_x, \mathbf{q}_c]) - \mathbf{q}_{obs}||^2 \ / \ 4\lambda^2)
\end{gathered}
\end{equation}

The push from a particular collision point, $\mathbf{q}_{obs}$, is given in the direction of $\mathbf{q}_{obs}$ towards its projection onto the candidate search-space region, $R[\mathbf{q}_x, \mathbf{q}_c]$.  If this projection point lies ``behind'' $\mathbf{q}_x$, \textit{i.e.}, $\texttt{proj\_scalar}(\mathbf{q}_{obs}, R[\mathbf{q}_x, \mathbf{q}_c]) < 0$, the collision point is considered already passed, and the $\psi_{3,1}$ term removes this collision point's effect on the gradient.  If $\mathbf{q}_{obs}$ lies ``ahead'' of $\mathbf{q}_x$, $\psi_{3,2}$ attenuates the push strength away from $\mathbf{q}_{obs}$ such that a smaller distance between $\mathbf{q}_{obs}$ and its projection point corresponds to a stronger push, and vice versa.  The scalar $5$ coefficient on the $\psi_{3,2}$ term raises the peak of its corresponding Gaussian to further strengthen the push away from $\mathbf{q}_{obs}$ if it is close to its projection onto $R[\mathbf{q}_x, \mathbf{q}_c]$.  The set of nearby collision states, $N_{obs}$, is efficiently accessed in constant time, \textit{i.e.}, without using a heavy data structure like a kd-tree, by using the checkpoint data back-propagation technique outlined in \S\ref{sec:local_planning_level} (Alg. \ref{alg:collision_points}).  Here, collision points are stored and accessed at checkpoint nodes in the $H_{obs}$ hashmap throughout Alg. \ref{alg:localsearch}.  Note that if $|N_{obs}| \equiv 0$, the effect of $\frac{\partial \textit{g}_3}{\partial \mathbf{q}_c}$ is cancelled by the Heaviside step coefficient in Alg. \ref{alg:local_edge}, line 6, and thus, $\frac{\partial \textit{g}_3}{\partial \mathbf{q}_c}$ does not need to be computed in this case.




The $\frac{\partial Pr_3(\ U_{\delta}(R[\mathbf{q}_x, \mathbf{q}_c])\ )}{\partial \mathbf{q}_c}$ gradient is used in gradient ascent iterations to steer the local tree, as seen in Alg. \ref{alg:local_edge}.  Because this gradient ascent occurs at such a performance critical inner-loop process, we only do one or two iterations per local search loop in practice.  By default, we use the point $\mathbf{q}_x + (\mathbf{q}_x - \mathbf{q}_p)$ as the initial condition for $\mathbf{q}_c$ in the gradient ascent.  Also, if $|N_{obs}| > 0$, we add a small amount of noise to this initial condition.  In our prototype system, this noise vector is drawn from a $d$-dimensional uniform distribution $\xi \sim \textit{U}^d(\frac{-\lambda}{100}, \frac{\lambda}{100})$.  Without this small random push, the local search would just stay on a single plane embedded in the configuration space.  We also re-set the edge length to $\lambda$ after each gradient ascent iteration (Alg. \ref{alg:local_edge}, line 7).


\algCollisionPoints

\section{Analysis}
\label{sec:analysis}
In this section, we show the probabilistic completeness property of SPRINT in $\mathbb{R}^d$.



\vspace{3pt}\noindent\textit{Sketch Proof}: Consider $\Gamma[\mathbf{q}_{init}, \mathbf{q}_{goal}]$ as any feasible solution path found by a probabilistically complete planner parameterized by a set of $k$ nodes, $N \equiv \{ \mathbf{q}_1, \mathbf{q}_2, ..., \mathbf{q}_k\}$, connected by linear edges, $E$.  For example, solution paths found by RRT or PRM would fit this definition.  Now, suppose $d$-dimensional open balls of radii $\textit{r}_1 \hdots \textit{r}_k$, are centered around nodes $\mathbf{q}_1 \hdots \mathbf{q}_k$, which we will denote as $\mathcal{B}_{\textit{r}_1}(\mathbf{q}_1) \hdots \mathcal{B}_{\textit{r}_k}(\mathbf{q}_k)$.  Each radius $\textit{r}_j$ will be selected such that all of $\mathcal{B}_{\textit{r}_j}(\mathbf{q}_j)$ lies in free-space and all points in $\mathcal{B}_{\textit{r}_{j}}(\mathbf{q}_{j})$ can be connected by collision-free lines with all points in $\mathcal{B}_{\textit{r}_{j+1}}(\mathbf{q}_{j+1}) \ \forall j \in [1, 2, ..., k-1]$.  Note that because $\Gamma[\mathbf{q}_{init}, \mathbf{q}_{goal}]$ is feasible, a radius of $\textit{r}_j > 0$ must exist for all $\mathcal{B}_{\textit{r}_{j}}(\mathbf{q}_{j})$  \cite{choset2005principles}.


\textbf{Observation 1}: Because milestone points are uniformly sampled from $\chi_{free}$ and each radius $\textit{r}_j > 0$, each $\mathcal{B}_{\textit{r}_j}(\mathbf{q}_{j})$ will become dense with milestones in the limit \cite{choset2005principles}.  \textbf{Observation 2}: SPRINT will process a local search between all current global node and milestone point pairs each loop through Algorithm \ref{alg:outerloop} lines 2--16.  \textbf{Observation 3}: If a collision-free, straight-line path exists between a global node $\mathbf{q}_n$ and a milestone $\mathbf{q}_m$, the local search in SPRINT resolves to a straight-line.  From these observations, it can be seen that the following recursive sequence will always happen in SPRINT: a milestone point will eventually be sampled in $\mathcal{B}_{\textit{r}_j}(\mathbf{q}_j)$, which will be reached by a straight line local search from a global node in $\mathcal{B}_{\textit{r}_{j-1}}(\mathbf{q}_{j-1}) \ \forall j \in [2, 3, ..., k]$.  Note that $\mathbf{q}_{init}$, \textit{i.e.}, $\mathbf{q}_1$, and $\mathbf{q}_{goal}$, \textit{i.e.}, $\mathbf{q}_k$, start as a global node and milestone point, respectively, so these points bootstrap the recursive process and never have to be exactly sampled.  Therefore, SPRINT must inherit the probabilistic completeness of any algorithm that would eventually find a feasible path $\Gamma[\mathbf{q}_{init}, \mathbf{q}_{goal}]$ as a homotopically equivalent path will always eventually be found by SPRINT. $\square$     

\section{Evaluation}
\label{sec:evaluation}

\begin{figure*}[t!]
	\includegraphics[width=\textwidth]{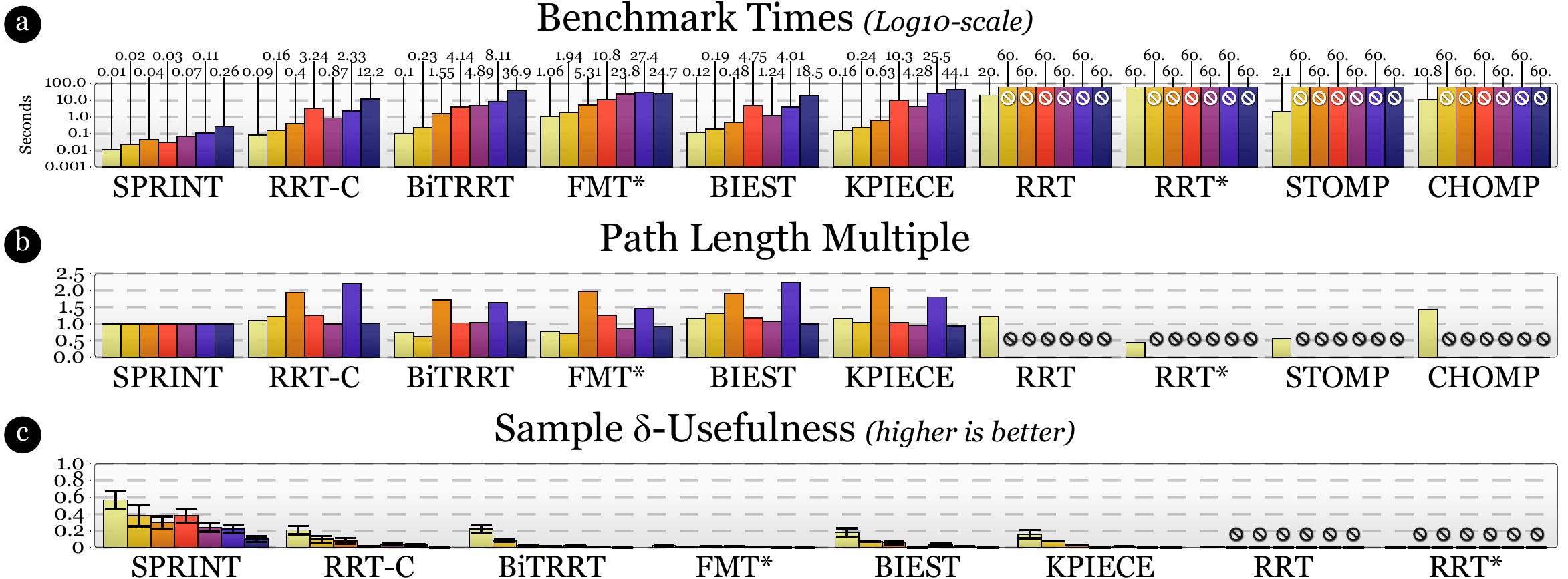}
	\caption{Results from Experiments 1 and 2.  Bar colors denote different tasks, with ordering from left to right being \textit{Single Box} (UR5), \textit{Table} (UR5), \textit{Vertical Bars} (Sawyer), \textit{Narrow Passage Box} (Sawyer), \textit{Over Boxes} (Hubo+), \textit{Bookshelf} (Hubo+), and \textit{Arms around Table} (Hubo+).  Error bars denote standard error.  The values above the log10-scale bars in (a) denote the standard decimal values of the bars (in seconds).  A circle with a line through it indicates that no solution was found on the given task in the allotted time.}  
	\label{fig:results_1}
	\vspace{-10pt}
\end{figure*}

We have assessed the efficacy of our approach in three empirical experiments. Although we demonstrate that SPRINT is able to solve low-dimensional testbed problems used to validate planning algorithms, as illustrated in Figure \ref{fig:teaser}, we focus our assessment on higher dimensional problems with robot manipulators.  In this section, we overview our experiments and share results.  

\subsection{Implementation Details}
\label{sec:implementation_details}
Our prototype SPRINT implementation is implemented in the Rust programming language.  Robot manipulator self-collisions and environment collisions are detected using the \texttt{ncollide} Rust library.  Configurations that exceed joint position limits are also considered in a collision state.  All evaluations throughout this work were run on a Lenovo Legion laptop with an i7-9750H processor and 32GB RAM.  All parameters used in SPRINT were held constant at their default values for all problems in our evaluation.


\subsection{Evaluation Benchmark}
We developed a set of seven benchmark tasks to compare our method against alternative path planners.  The benchmark was designed to test the planners on a wide variety of tasks that range in dimensionality and topological structure.  Our benchmark consists of two tasks on a Universal Robots UR5 (6-DOF), two tasks on a Rethink Robotics Sawyer (7-DOF), and three bimanual tasks on the DRC-Hubo+ (15-DOF).  Two of these tasks can be seen illustrated in Figure \ref{fig:teaser}(b--c).  The UR5 tasks included (1) \textit{Single Box}, where the robot moves over a large box; and (2) \textit{Table}, where the robot moves its arm up and around a small table.  The Sawyer tasks included (3) \textit{Vertical Bars} where the robot maneuvers around six vertical bars; and (4) \textit{Narrow Passage Box} where the robot maneuvers out of a cramped box through a narrow passage.  Finally, the Hubo tasks included (5) \textit{Over Boxes} where the robot moves its arm up and over a set of boxes; (6) \textit{Bookshelf} where the robot moves an item from the top shelf to the bottom shelf; and (7) \textit{Arms around Table} where the robot maneuvers both of its arms up and around a table to reach objects on top.  Our benchmark consisted of 100 trials through each task.  All tasks had a maximum evaluation time of one minute.




\subsection{Experiment 1: Comparisons with Alternative Planners}
In our first experiment, we compared performance on our benchmark tasks to nine commonly used path planners: RRT \cite{lavalle2001randomized}, RRT-Connect \cite{kuffner2000rrt}, RRT$^*$\cite{karaman2011sampling}, FMT$^*$ \cite{janson2015fast}, KPIECE \cite{csucan2009kinodynamic}, BiEST \cite{hsu1997path}, BiTRRT \cite{jaillet2010sampling}, CHOMP \cite{ratliff2009chomp}, and STOMP \cite{kalakrishnan2011stomp}.  All methods were tested using their MoveIt! implementations with default planner options.  Also, we implemented RRT, RRT-Connect, FMT$^*$, and EST within our framework to ensure that the planning primitives in SPRINT, \textit{e.g.}, collision checking or graph operations, were not eliciting unintended performance gains.  Our implementations used kd-trees for nearest neighbor checking.  We used both our own implementations and MoveIt! in our evaluation and only report the better of the two results for fairness.  Across all planners and tasks, the highly optimized implementations in MoveIt! were faster.  



\subsubsection{Results---Computation Time}
Figure \ref{fig:results_1}a provides an overview of the average times needed to complete our benchmark tasks.  SPRINT computed solution paths for the benchmark problems often one to three orders of magnitude faster than alternative approaches.  While many of the planners did well on the somewhat simpler, lower dimensional UR5 problems, SPRINT still out-performed the other methods on these tasks.  


\subsubsection{Results---Path Length}
Figure \ref{fig:results_1}b provides an overview of average path lengths per each planner and task.  The paths computed by SPRINT are shorter or at most similar in length compared to other sampling-based planners.  Thus, the performance advantages in computation time do not come at the expense of lower quality paths.  While optimizing planners, such as RRT$^*$ and STOMP, did find higher quality paths on the UR5 \textit{Single Box} task, these results came at a significantly higher computation cost.  Further, these planners were not able to find any solutions in the allotted sixty seconds for any of the other benchmark tasks.  

\subsection{Experiment 2: Analysis of Performance Gains}
In our second experiment, our goal was to investigate whether the performance gains exhibited by SPRINT in Experiment 1 could be attributed to enhanced sample efficiency.  We assessed the ratio of points from the solution paths from Experiment 1 that were $\delta-$Useful for each tested planner, which we defined in \S\ref{sec:search_space_regions_useful_samples}.  We defined $\delta$ to be $2\lambda$ for this evaluation, where $\lambda$ was the minimum edge length in the tree or graph structure.  We only report this $\delta$-Usefulness metric on the sampling-based planners as it is not applicable to the local-optimization-based planners (STOMP and CHOMP).



\begin{figure}[t!]
	\includegraphics[width=\columnwidth]{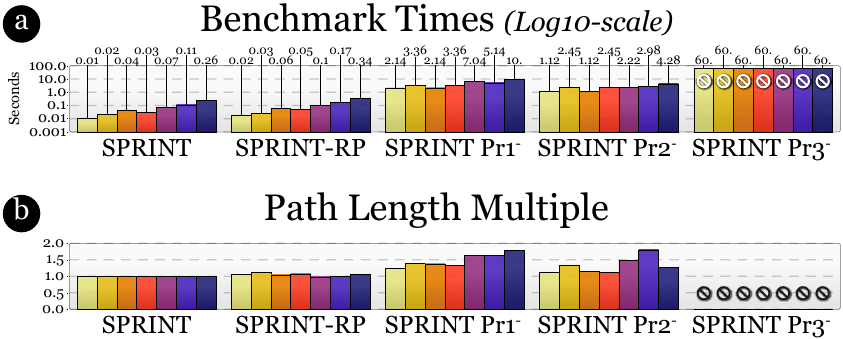}
	\caption{Results from Experiment 3.  Bar colors denote different tasks, with ordering from left to right indicating \textit{Single Box} (UR5), \textit{Table} (UR5), \textit{Vertical Bars} (Sawyer), \textit{Narrow Passage Box} (Sawyer), \textit{Over Boxes} (Hubo+), \textit{Bookshelf} (Hubo+), and \textit{Arms around Table} (Hubo+).  The values above the log10-scale bars in (a) denote the standard decimal values of the bars (in seconds).  A circle with a line indicates that no solution was found.}  
	\label{fig:results_3}
	\vspace{0pt}
\end{figure}

Figure \ref{fig:results_1}c shows the $\delta-$Useful metric results.  We observe that SPRINT performs much better on this metric across all tasks, indicating that many more of its samples were useful.  These results suggest that the performance gains from SPRINT were indeed due to our intended goal of enhancing sample-efficiency.  


\subsection{Experiment 3: Analysis of Heuristics and Parameters}
In our final experiment, our goal was to observe how the performance of SPRINT changes if we systematically perturb its parameters or remove its probability heuristics.  We assessed four conditions: (1) SPRINT-RP, which randomly offsets all parameters such that each varies by up to 25\% above or below its original hand selected value.  These random offsets were drawn from a uniform distribution; SPRINT $Pr_1^-$, which replaced the $Pr_1$ heuristic with a random global edge selection; (2) SPRINT $Pr_2^-$, which replaced the $Pr_2$ heuristic with a random, $50\%$ chance of extending a node; and (4) SPRINT $Pr_3^-$, which replaced the $Pr_3$ heuristic with a random extend direction process.


Figure \ref{fig:results_3} overviews the results from Experiment 3.  First, we found that SPRINT-RP was comparable in terms of computation time and average path length, suggesting that SPRINT is not overly sensitive to parameter tuning.  However, we see that performance significantly degrades when even one probability heuristic is removed, suggesting that all of our probability heuristics are integral and work in tandem in order to enhance sample-efficiency and, in turn, boost performance.  Probability Heuristic 3 appears to be particularly important, as no solutions were found in the allotted sixty seconds without this heuristic model.   





\section{Discussion}
\label{sec:discussion}
In this work, we presented a path planning approach that is able to quickly and reliably solve high-dimensional path planning problems.  Through the notion of enhanced sample usefulness afforded by a set of probability heuristics, we show that our approach achieves significant performance gains over standard approaches.


\textit{Limitations}---We note a number of limitations of our work that suggest future extensions.  First, our probability heuristics were constructed by observation and intuition.  While our three heuristics serve as proofs of concept for our overall premise, we believe that better models, either hand-engineered or data-driven, could improve results.  Our approach also does not guarantee asymptotic optimality.  Lastly, our work does not currently accommodate kinodynamic planning, though we believe that extensions to our work could explore kinodynamic steering functions compatible with Probability Heuristic 3.

\section*{ACKNOWLEDGMENTS}
The authors thank Lydia Kavraki for valuable discussions regarding this work.


\bibliographystyle{IEEEtran}
\bibliography{references}

\end{document}